\documentclass[letterpaper, 10 pt, journal, twoside]{IEEEtran}
\usepackage{amsmath,amsfonts,amssymb}
\usepackage{bm}
\usepackage{algorithm}% http://ctan.org/pkg/algorithms
\usepackage{algpseudocode}% http://ctan.org/pkg/algorithmicx
\usepackage{array}
\usepackage[caption=false,font=normalsize,labelfont=sf,textfont=sf]{subfig}
\usepackage{textcomp}
\usepackage{stfloats}
\usepackage{url}
\usepackage{verbatim}
\usepackage{graphicx}

\usepackage[T1]{fontenc} % for Berner2019a
\usepackage{color} % used for the command \revised

\newcommand{\bx}{\bm{x}}
\newcommand{\bu}{\bm{u}}

\algnewcommand{\LeftComment}[1]{\Statex \hspace{4mm} \(\triangleright\) #1}

\newcommand\revised[2]{\textcolor{#2}{#1}}  % used for the revision

\def\BibTeX{{\rm B\kern-.05em{\sc i\kern-.025em b}\kern-.08em
    T\kern-.1667em\lower.7ex\hbox{E}\kern-.125emX}}
\usepackage{balance}

\begin{document}

\markboth{IEEE Robotics and Automation Letters. Preprint Version.
  Accepted July, 2022}
{Uchibe: Model-Based Imitation Learning Using Entropy Regularization of
  Model and Policy} 
% Use only for final RAL version

\title{Model-Based Imitation Learning Using Entropy Regularization of
  Model and Policy}

\author{Eiji Uchibe$^{1}$% <-this % stops a space
  \thanks{Manuscript received: February, 24, 2022;
    Revised May, 20, 2022;
    Accepted July, 11, 2022.}%Use only for final RAL version
  \thanks{This paper was recommended for publication by
    Associate Editor J. Garcia and Editor D. Kulic upon evaluation
    of the reviewers' comments.
    This work was supported by Innovative Science and Technology
    Initiative for Security Grant Number JPJ004596, ATLA,
    Japan and partially based on results obtained from project
    JPNP20006 commissioned by the New Energy and Industrial
    Technology Development Organization (NEDO). This work
    was partially supported by JSPS KAKENHI Grant Number
    JP21H03527.}% <-this % stops a space
  \thanks{$^{1}$Eiji Uchibe is with the Department of Brain Robot
    Interface, ATR Computational Neuroscience Laboratories,
    Kyoto 619-0288, Japan. {\tt\footnotesize uchibe@atr.jp}}%
  \thanks{Digital Object Identifier (DOI): see top of this page.}
}

\maketitle

\begin{abstract}
  Approaches based on generative adversarial networks for imitation
  learning are promising because they are sample efficient in terms
  of expert demonstrations.
  However, training a generator requires many interactions with the
  actual environment because model-free reinforcement learning is
  adopted to update a policy.
  To improve the sample efficiency using model-based reinforcement
  learning, we propose model-based Entropy-Regularized Imitation
  Learning (MB-ERIL) under the entropy-regularized Markov decision
  process to reduce the number of interactions with the actual
  environment.
  MB-ERIL uses two discriminators. A policy discriminator
  distinguishes the actions generated by a robot from expert ones,
  and a model discriminator distinguishes the counterfactual state transitions
  generated by the model from the actual ones.
  We derive structured discriminators so that the learning of the
  policy and the model is efficient.
  Computer simulations and real robot experiments show that MB-ERIL
  achieves a competitive performance and significantly improves
  the sample efficiency compared to baseline methods. 
\end{abstract}


\section{INTRODUCTION}
\label{sec:introduction}

\IEEEPARstart{D}{eep} Reinforcement Learning (RL) using
deep neural networks learns much better than manually
designed policies (action rules) for problems where the
environmental dynamics is completely described on a
computer. 
Unfortunately, many problems remain in applying RL to robot
control tasks, especially two significant ones:
(1) specifying reward functions and (2) the cost of collecting
data in real environments.

\begin{figure}[t]
  \centering
  \includegraphics[width=1.0\hsize]{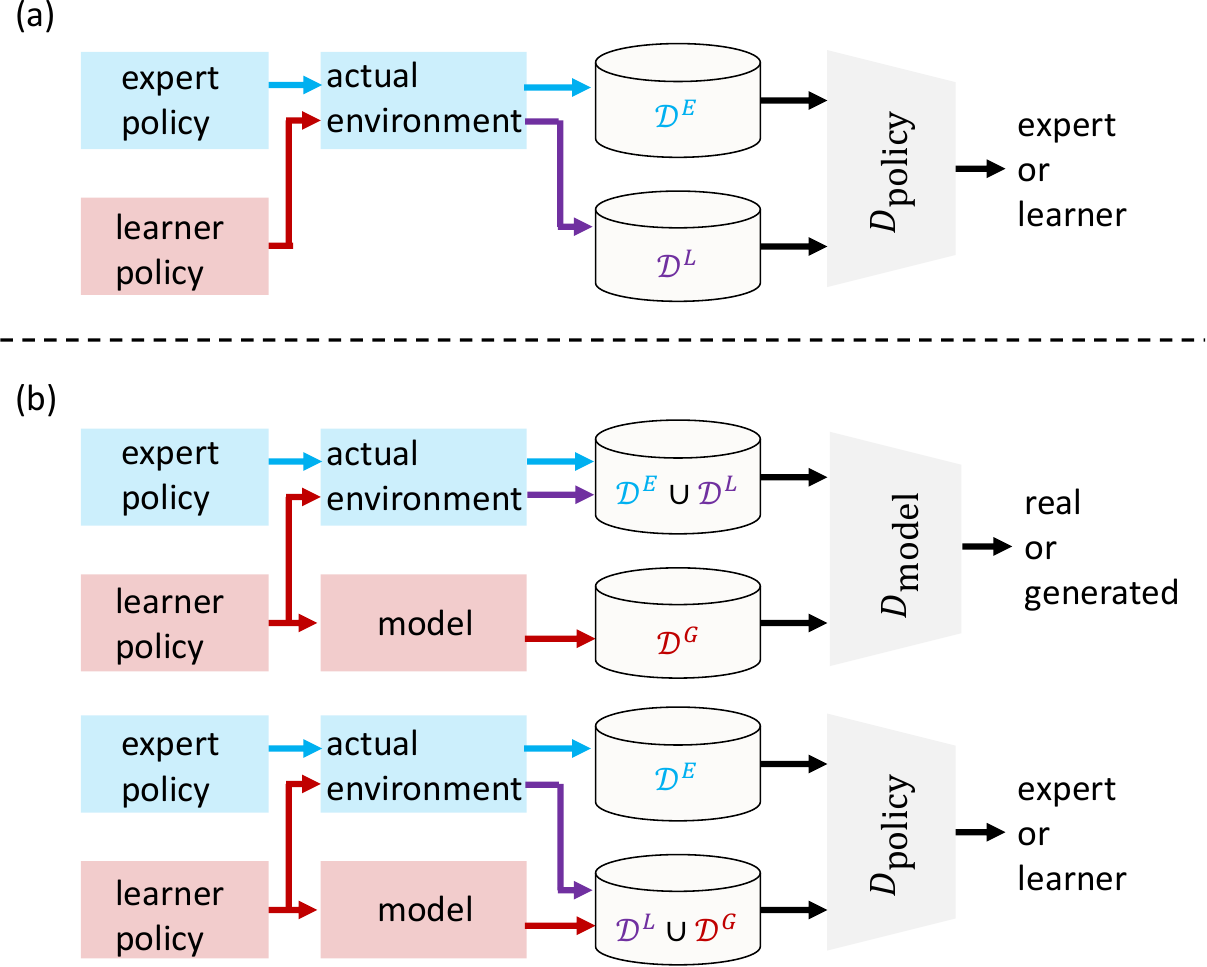}
  \caption{Comparison of available datasets and discriminators
    between model-free and model-based imitation learning:
    (a) Model-free setting: Expert dataset
    $\mathcal{D}^E$ and learner's dataset $\mathcal{D}^L$ are
    created through interactions with actual environments.
    Policy discriminator $D_{\mathrm{policy}}$ exists.
    (b) Our proposed model-based setting: Simulated dataset
    $\mathcal{D}^G$ is additionally generated from learner's
    policy and model. Two discriminators exist:
    $D_{\mathrm{model}}$ and $D_{\mathrm{policy}}$.}
  \label{fig:discriminators}
\end{figure}

Imitation learning is a promising method for overcoming the first
problem because it finds a policy from expert demonstrations.
In particular, recent imitation learning is related to Generative
Adversarial Networks (GANs) \cite{Finn2016d, Ho2016c}, and
this framework has two components: a discriminator and
a generator.
Training the former, which distinguishes demonstrations
generated by a robot from expert demonstrations, corresponds to inverse
RL \cite{AbAza2020a, Arora2021a}.
Training the latter, which produces expert-like demonstrations,
corresponds to policy improvement by forward RL\footnote{Hereafter
  we refer to RL that finds an optimal
  policy from rewards as forward RL to distinguish it from
  inverse RL.}.
Some previous studies \cite{Ho2016c, Fu2018a}
showed that such imitation learning approaches achieve high sample
efficiency in terms of the number of demonstrations.
However, they also often suffer from sample
inefficiency in the generator's training step because a
model-free on-policy forward RL, which usually requires many
environmental interactions, is adopted to update the policy.
Several studies \cite{Blonde2019a, Kostrikov2019a, Sasaki2019a,
  Uchibe2021a} employed model-free off-policy RL. 
However, these methods remain sample inefficient as a
method of real robot control because they require costly
interactions with actual environments. 
Fig.~\ref{fig:discriminators}(a) illustrates the dataset and
the discriminator in the model-free setting.
Expert dataset $\mathcal{D}^E$ is gathered by executing
an expert policy in an actual environment.
Note that the learner also interacts with
the actual environment using its own policy to collect 
learner's dataset $\mathcal{D}^G$, although its interaction
is costly. 
Adopting a model-based forward RL is promising to reduce the
number of costly interactions with the actual environment.

To further reduce the number of interactions with the actual
environment, we propose
Model-Based Entropy-Regularized Imitation Learning (MB-ERIL), which
explicitly estimates a model of the environment and 
generates simulated data by running the learner's policy in
the estimated model.
MB-ERIL is formulated as the minimization problem of
Kullback-Leibler (KL) divergence between the expert policy
evaluated in the actual environment and the learner's policy
evaluated in the model environment.
MB-ERIL regularizes the policy and model by Shannon entropy
and KL divergence to derive the algorithm, where we assume that
the expert policy and the actual environment are solutions of
the regularized Bellman equation.

The following are the contributions of this work:
\begin{itemize}
\item It develops two novel discriminators.
  One is a policy discriminator that differentiates actions
  generated by the policy from expert actions.
  The other is that a model discriminator that distinguishes the
  state transitions generated by the model from the actual
  transitions provided by an expert.
\item The discriminators are represented by the reward, state,
  and state-action value functions. The value functions are
  updated by training the discriminators, a step that makes the
  forward RL efficient.
\item For model fitting, MB-ERIL provides another objective
  alternative to maximum likelihood estimation.
\item The estimated model generates counterfactual data to
  update the state and state-action value functions
  (Fig.~\ref{fig:discriminators}(b)).
  As a result, we can reduce the number of costly interactions
  with the actual environment.
\end{itemize}

To evaluate MB-ERIL, we conducted two continuous control benchmark
tasks in the MuJoCo simulator \cite{Todorov2012a} and a
vision-based reaching task \cite{Uchibe2021a} using a real
upper-body humanoid robot.
MB-ERIL shows promising experimental results with competitive
asymptotic performance and higher sample efficiency than
previous studies and
 empirically ensures low model bias, which is the gap between the
actual environment and the model, by jointly updating the
model and policy.

\section{LITERATURE REVIEW}
\label{sec:literature_review}

\subsection{Model-free imitation learning}
\label{sec:model_free_IL}

Relative Entropy Inverse RL \cite{Boularias2011a} is a
sample-based method inspired by Maximum entropy inverse
RL (MaxEntIRL) \cite{Ziebart2008a}. It provides an
efficient way of estimating the partition function of the
probability distribution of expert trajectories.
However, one drawback is utilizing a fixed sampling
distribution, which is unhelpful in practice.
Generative Adversarial Imitation Learning (GAIL) \cite{Ho2016c}, which
can adapt the sampling distribution using policy optimization,
is closely related to GAN. GAIL is more sample efficient
than Behavior Cloning (BC) with respect to the number of
expert demonstrations.
However, GAIL often requires many interactions
with the environment because (1) an on-policy model-free forward RL
is used for policy improvement and (2) the discriminator is
not structured.
Adversarial Inverse Reinforcement Learning (AIRL) \cite{Fu2018a}
and Logistic Regression-based inverse RL 
\cite{Uchibe2018c} propose a structured discriminator
represented by the reward and state-value functions,
although an on-policy forward RL updates the policy.
Adversarial Soft Advantage Fitting (ASAF) \cite{Barde2020a}
exploits the discriminator inspired by the AIRL discriminator,
although the state transition is not considered explicitly.
To reduce the number of environmental interactions, an off-policy
model-free forward RL has been adopted
\cite{Blonde2019a, Kostrikov2019a, Zuo2020a, Zuo2020b}.
We proposed Model-Free Entropy-Regularized Imitation
Learning (MF-ERIL) based on entropy regularization that
shares the network parameters between the discriminator and the
generator.
Even though MF-ERIL achieved better sample efficiency than the above methods,
there is room to improve the sample efficiency by introducing
model learning.
MB-ERIL is one instantiation of this approach.
Recent approaches corrected the issues around reusing data previously collected
while training the discriminator \cite{Zhu2020a, Hoshino2022a}.

GAN-like imitation learning often suffers from unstable learning
from adversarial training.
Therefore, some regularization terms are added to BC's loss function
to stabilize the learning process.
For example, Soft Q Imitation Learning (SQIL) is a regularized
BC algorithm, where the squared soft Bellman error
is used as a regularizer \cite{Reddy2020a}.
Interestingly, its algorithm can be implemented by assigning a
reward of 1 to expert demonstrations and 0 to generated ones.
Discriminator Soft Actor-Critic \cite{Nishio2020a} is an extension
of SQIL where its predefined reward value is replaced
by the parameterized reward trained by the discriminator.

\subsection{Model-based imitation learning}
\label{sec:model_based_IRL}

MaxEntIRL, which is a pioneer of an entropy
regularized RL, estimates the reward function from expert
trajectories based on the maximum entropy principle.
MaxEnt IRL is a model-based approach, although 
how to train the model was not discussed.
Although a method was proposed that simultaneously estimated the reward and the
model \cite{Herman2016a}, maximum likelihood
estimation was applied to model learning.

A few studies introduced model learning to GAIL
to propagate the discriminator's gradient to the
policy for a gradient update.
Consequently, Model-based GAIL \cite{Baram2017a, Rhinehart2018a}
and Model-based AIRL \cite{Sun2021a} achieved end-to-end training,
although these methods do not sample state transitions from the model
while unrolling the trajectories.
However, these approaches do not use simulated experiences through
interaction with the model, and therefore, they still require
many actual interactions with a real environment \cite{Saxena2020a}.
Data-Efficient Adversarial Learning for Imitation from Observation
(DEALIO), which is a GAIL-like algorithm \cite{Torabi2021a}, 
adopts a model-based RL that is employed for training a policy
from the trained discriminator.
Since one drawback of DEALIO is that model learning is independent of
discriminator training, it suffers from 
model bias.

\section{PROPOSED METHOD}
\label{sec:proposed_method}

\subsection{Objective function}
\label{sec:objective_function}

Consider Markov Decision Process (MDP)
$\mathcal{M} = \langle \mathcal{X}, \mathcal{U}, p,
r, \gamma \rangle$, where $\mathcal{X}$ denotes a continuous state
space, $\mathcal{U}$ denotes an action space,
$p: \mathcal{X} \times \mathcal{U} \times \mathcal{X} \mapsto
\mathbb{R}_{\geq 0}$ denotes an actual stochastic state transition probability,
$r: \mathcal{X} \mapsto \mathbb{R}$ denotes an immediate reward
function, and $\gamma \in (0, 1)$ denotes a discount factor
that indicates how near and far future rewards are weighed.
We chose the state-only function because more
general reward functions like $r: \mathcal{X} \times \mathcal{U}
\times \mathcal{X} \mapsto \mathbb{R}$ are often shaped
as the result of training the discriminators. 
Let $b: \mathcal{X} \times \mathcal{U} \mapsto
\mathbb{R}_{\geq 0}$ denote a stochastic policy to select
action $\bu \in \mathcal{U}$ at state $\bx \in \mathcal{X}$.
Let $\pi$ and $q$ denote an expert policy and
a model of $p$.  Note that $p$ and $\pi$ are unknown,
while $q$ and $b$ are maintained by the learner.

MB-ERIL minimizes the following KL divergence:
\begin{displaymath}
  J(q, b) = \mathbb{E}_{p^L(\bx, \bu, \bx')} \left[ \ln \frac{p^L (\bx, \bu, \bx')}
  {p^E (\bx, \bu, \bx')} \right],
\end{displaymath}
where $p^L$ and $p^E$ are respectively the joint density functions
defined by
\begin{align}
  p^L(\bx, \bu, \bx') &= q(\bx' \mid \bx, \bu) b(\bu \mid \bx) p^L(\bx),
  \notag \\
  p^E(\bx, \bu, \bx') &= p(\bx' \mid \bx, \bu) \pi(\bu \mid \bx) p^E(\bx),
  \notag 
\end{align}
where $p^L(\bx)$ and $p^E(\bx)$ are some initial distributions, and
we assume $p^L(\bx) = p^E(\bx)$ for simplicity.
The difficulty is how to evaluate the log-ratio because it is unknown.  

\subsection{Entropic regularization of policy and model}
\label{sec:entropic_regularization}

The basic idea for estimating the log-ratio is to adopt the density ratio
trick \cite{Sugiyama2012b}.
Then MB-ERIL updates the policy and the model by minimizing
the estimated KL divergence. 
To derive the algorithm, we formulate the entropic regularization of the
policy and the model. We add two regularization terms to the immediate
reward:
\begin{align}
  & r(\bx) + \kappa^{-1} 
  \mathcal{H}(p(\cdot \mid \bx, \bu) \pi(\cdot \mid \bx))
  \notag \\
  & \qquad
  - \eta^{-1} \mathrm{KL}(p(\cdot \mid \bx, \bu) \pi(\cdot \mid \bx)
  \parallel q(\cdot \mid \bx, \bu) b(\cdot \mid \bx)),
  %% \label{eq:entropy_regularized_reward}
  \notag 
\end{align}
where
$\mathcal{H}$ represents the operator of Shannon entropy,
and $\kappa$ and $\eta$ denote positive hyperparameters.

Then the Bellman optimality equation is given by
\begin{align}
  & V(\bx) = \max_{\pi} \mathbb{E}_{\pi} \left\{
    - \kappa^{-1} \ln \pi (\bu \mid \bx) 
    - \eta^{-1} \ln \frac{\pi (\bu \mid \bx)}{b (\bu \mid \bx)}
    \right. \notag \\
    &\quad + \max_{p} \mathbb{E}_{p} \biggl[
        r(\bx) + \gamma V(\bx') 
        \notag \\
    &\quad \quad 
    \left.
        - \kappa^{-1} \ln p (\bx' \mid \bx, \bu) 
        - \eta^{-1} \ln \frac{p (\bx' \mid \bx, \bu)}
                             {q (\bx '\mid \bx, \bu)}
    \biggr] \right\},
    \label{eq:Bellman_equation}
\end{align}
where $V(\bx)$ denotes a state-value function. 
To utilize the framework of entropic regularization, 
MB-ERIL assumes that $\pi(\bu \mid \bx)$ and
$p(\bx' \mid \bx, \bu)$ are the solutions of the
Bellman equation \eqref{eq:Bellman_equation} when the learner
uses baseline policy $b(\bu \mid \bx)$ and baseline state
transition $q(\bx' \mid \bx, \bu)$.
Using the Lagrangian multiplier method, we obtain the following
equations:
\begin{align}
  & p(\bx' \mid \bx, \bu) \notag \\
  &\quad =
    \frac{\exp [\beta (r(\bx) + \gamma V(\bx') + \eta^{-1}
    \ln q(\bx' \mid \bx, \bu))]}
    {\exp [\beta Q(\bx, \bu)]},
  \label{eq:model_update_rule}
\end{align}
\begin{align}
  & \pi (\bu \mid \bx) =
    \frac{\exp [\beta (Q(\bx, \bu) + \eta^{-1} \ln b(\bu \mid \bx))]}
    {\exp (\beta V(\bx))},
  \label{eq:policy_update_rule}
\end{align}
where $\beta$ is a positive hyperparameter defined by
$\beta \triangleq \frac{\kappa \eta}{\kappa + \eta}$.
$Q(\bx, \bu)$ denotes the state-action value function, and
the following relation exists between $V$ and $Q$:
\begin{multline}
  \exp ( \beta Q(\bx, \bu) ) = \int \exp \left[
    \beta ( r(\bx) + \gamma V(\bx') \right. \\
     \left. + \eta^{-1} \ln q
    (\bx' \mid \bx, \bu) 
  \right] \mathrm{d}\bx',
  \label{eq:V_and_Q1}
\end{multline}
\begin{equation}
  \exp (\beta V(\bx)) = \int \exp \left[\beta (Q(\bx, \bu) + \eta^{-1}
  \ln b(\bu \mid \bx)) \right] \mathrm{d}\bu.
  \label{eq:V_and_Q2}
\end{equation}
See Appendix~A for the derivation.

\subsection{Derivation of MB-ERIL discriminators}
\label{sec:discriminators_MB-ERIL}

We obtain the following equations from
\eqref{eq:model_update_rule} and \eqref{eq:policy_update_rule}
that represent the density ratio:
\begin{align}
  \beta^{-1} \ln \frac{p (\bx' \mid \bx, \bu)}
  {q(\bx' \mid \bx, \bu)} &= r(\bx) + \gamma V(\bx')
    - Q(\bx, \bu) \notag \\
    &\quad - \kappa^{-1} \ln q(\bx' \mid \bx, \bu),
    \notag
\end{align}
\begin{align}
  \beta^{-1} \ln \frac{\pi (\bu \mid \bx)}
  {b(\bu \mid \bx)} = Q(\bx, \bu) - V(\bx)
  - \kappa^{-1} \ln b(\bu \mid \bx).
    \notag
\end{align}

Using the density ratio estimation \cite{Sugiyama2012b},
we derive a model discriminator that distinguishes the state
transition generated by the model from the real transitions
provided by the expert:
\begin{align}
  & D_{\mathrm{model}} (\bx' \mid \bx, \bu) \notag \\
  &\quad =
  \frac{\exp (\beta f(\bx, \bu, \bx'))}
  {\exp (\beta f(\bx, \bu, \bx')) + \exp(\beta \kappa^{-1}
  \ln q(\bx' \mid \bx, \bu) )},
  \notag
\end{align}
where $f(\bx, \bu, \bx')$ is defined by
\begin{displaymath}
  f(\bx, \bu, \bx') \triangleq
  r(\bx) + \gamma V(\bx') - Q(\bx, \bu).
\end{displaymath}
Similarly, the policy discriminator is given by
\begin{align}
  & D_{\mathrm{policy}} (\bu \mid \bx) \notag \\
  &\quad =
  \frac{\exp (\beta (Q(\bx, \bu) - V(\bx)))}
    {\exp (\beta (Q(\bx, \bu) - V(\bx))) +
    \exp (\beta \kappa^{-1} \ln b(\bu \mid \bx))},
    \notag
\end{align}
which differentiates the action generated by
the policy from the expert action.
Note that $D_{\mathrm{policy}}$ is the discriminator used by
AIRL and MF-ERIL if $Q(\bx, \bu)$ is replaced with
$r(\bx) + \gamma V(\bx')$. 
These discriminators have the same form as the optimal
discriminator, also known as the Bradley-Terry model
\cite{Bradley1952a}. 

\subsection{MB-ERIL learning}
\label{sec:learning_MB-ERIL}

In the model-free setting, the learner needs to interact
with a real environment to generate trajectories like
the expert, making the model-free setting sample
inefficient.
On the other hand, the learner can generate trajectories
by interacting with the real environment and the model.
Learning from simulated trajectories is sometimes called
Dyna architecture \cite{Sutton1990a}.
To improve the sample efficiency, we adopt Dyna
architecture, and therefore, MB-ERIL utilizes the following
three datasets.
The first is an expert dataset generated by an expert
in a real environment:
\begin{align}
  & \mathcal{D}^E = \{ (\bx_i, \bu_i, \bx_i') \}_{i=1}^{N^E}, 
  \quad
  \bx_i \sim p_0^E (\cdot)
  \quad
  \notag \\  
  &\quad \bu_i \sim \pi (\cdot \mid \bx_i),
  \quad
  \bx_i' \sim p(\cdot \mid \bx_i, \bu_i),
  \notag
\end{align}
where $N^E$ is the number of transitions in the dataset and
$p_0^E$ is the discounted state distribution \cite{Sutton2000a}
for ($\pi$, $p$).
The second and third are the learner's datasets by
running $b(\bu \mid \bx)$ in the real environment and the model:
\begin{align}
  & \mathcal{D}^L = \{ (\bx_i, \bu_i, \bx_i') \}_{i=1}^{N^L}, 
  \quad
  \bx_i \sim p_0^L (\cdot)
  \quad
  \notag \\  
  &\quad \bu_i \sim b (\cdot \mid \bx_i),
  \quad
  \bx_i' \sim p(\cdot \mid \bx_i, \bu_i),
  \notag \\
  & \mathcal{D}^G = \{ (\bx_i, \bu_i, \bx_i') \}_{i=1}^{N^G}, 
  \quad
  \bx_i \sim q_0 (\cdot)
  \quad
  \notag \\  
  &\quad \bu_i \sim b (\cdot \mid \bx_i),
  \quad
  \bx_i' \sim q(\cdot \mid \bx_i, \bu_i),
  \notag
\end{align}
where $N^L$ and $N^G$ are the dataset sizes, and $p_0^L$
and $q_0$ are the discounted state distributions for
($\pi^L$, $p$) and ($\pi^L$, $q$).
Normally, $N^G \gg N^L \geq N^E$, because collecting data in the real
environment is expensive.

Here we show MB-ERIL's learning rules that consist of (1)
training the discriminators, (2) policy evaluation, and (3) 
policy improvement.
First, we show the loss functions for training
$\mathcal{D}_{\mathrm{model}}$ and $\mathcal{D}_{\mathrm{policy}}$
to estimate $r$, $V$, and $Q$:
\begin{align}
  &\mathcal{L}_{\mathrm{model}} (r, V, Q) 
  = -
  \mathbb{E}_{(\bx, \bu, \bx') \sim \mathcal{D}^E \cup \mathcal{D}^L}
  \left[ \ln D_{\mathrm{model}} (\bx' \mid \bx, \bu) \right]
  \notag \\
  &\quad
  - \mathbb{E}_{(\bx, \bu, \bx') \sim \mathcal{D}^G}
  \left[ \ln (1 - D_{\mathrm{model}} (\bx' \mid \bx, \bu)) \right]
  \notag % \label{eq:loss_D_model},
\end{align}
where $\mathbb{E}_{(\bx, \bu, \bx' \sim \mathcal{D})}$ represents
the expectation over a batch of data that is uniformly sampled from an experience
replay buffer $\mathcal{D}$.
Similarly, the loss function for $D_{\mathrm{policy}}$ is
given by
\begin{align}
  &\mathcal{L}_{\mathrm{policy}} (V, Q) 
  = -
  \mathbb{E}_{(\bx, \bu) \sim \mathcal{D}^E}
  \left[ \ln D_{\mathrm{policy}} (\bu \mid \bx) \right]
  \notag \\
  &\quad 
  - \mathbb{E}_{(\bx, \bu) \sim \mathcal{D}^L \cup \mathcal{D}^G}
  \left[ \ln (1 - D_{\mathrm{policy}} (\bu \mid \bx)) \right].
  \notag % \label{eq:loss_D_policy}
\end{align}
Note that $r$, $V$, and $Q$ are all updated, and $q$ and $b$
remain fixed while training the discriminators.
These two loss functions are summarized:
\begin{multline}
  \mathcal{L}_{\mathrm{dis}}(r, V, Q) =
  \lambda_{\mathrm{model}}
  \mathcal{L}_{\mathrm{model}}(r, V, Q) \\ +
  \lambda_{\mathrm{policy}} \mathcal{L}_{\mathrm{policy}}(V, Q),
  \label{eq:loss_D}
\end{multline}
where $\lambda_{\mathrm{model}}$ and $\lambda_{\mathrm{policy}}$
denote positive hyperparameters. 
Training the discriminators is interpreted by an inverse RL because the
reward function is estimated.

Next we show how the state and state-action value functions are
updated. This corresponds to the policy evaluation step.
Using \eqref{eq:V_and_Q1} and \eqref{eq:V_and_Q2}, the
following loss function can be constructed:
\begin{align}
  & \mathcal{L} (V, Q) = \lambda_{QV} \mathbb{E}_{\bx, \bu} 
  \biggl[ \notag \\
  &\; \; 
  \left( Q - \beta^{-1} \ln
    \mathbb{E}_{q} \left[ \exp (\beta (r + \gamma V'
    - \kappa^{-1} \ln q)) \right]
  \right)^2        
  \biggr]
  \notag \\
  &\quad + \lambda_{VQ}
  \mathbb{E}_{\bx} \left[ \left( V - \beta^{-1} \ln
    \mathbb{E}_{b} \left[ \exp (\beta (Q 
    - \kappa^{-1} \ln b)) \right]
  \right)^2        
  \right],
  \label{eq:VQ_update_rule}
\end{align}
where $\lambda_{QV}$ and $\lambda_{VQ}$ denote 
constant hyperparameters. 
We simplify the notation by omitting the arguments of
the functions and setting $V' \triangleq V(\bx')$
for readability.
This loss function contains the log-expected-exponent terms,
which introduce biases in their gradients \cite{Kostrikov2020a}.
Empirically, using a biased estimate was sufficient for our
experiments, although applying Fenchel conjugates is also promising.

Finally, we explain how the model and policy are updated based on
the trained discriminators in the policy improvement step.
As Soft Actor-Critic does
\cite{Haarnoja2018a}, they are directly learned by
minimizing the expected KL divergence based on
\eqref{eq:model_update_rule} and \eqref{eq:policy_update_rule}:
\begin{align}
  & q(\bx' \mid \bx, \bu) \leftarrow \mathrm{arg} \min_{q'}
    \notag \\
  &\; \; \mathbb{E}_{\bx, \bu} \left[
    \mathrm{KL} \left( q' \parallel
    \frac{\exp [\beta (r + \gamma V' + \eta^{-1}
    \ln q)]}{\exp [\beta Q]} \right)
    \right],
  \label{eq:loss_model}
\end{align}
\begin{align}
  & b(\bu \mid \bx) \leftarrow \mathrm{arg} \min_{b'}
    \mathbb{E}_{\bx} \left[
    \mathrm{KL} \left( b' \parallel
    \frac{\exp [\beta (Q + \eta^{-1} \ln b)]}{\exp (\beta V)}
    \right)
    \right].
  \label{eq:loss_policy}
\end{align}
MB-ERIL is summarized in Algorithm~\ref{alg:MB-ERIL}.
Lines 7 and 8 correspond to learning from simulated trajectories
like Dyna architecture.

\begin{algorithm}[t]
\caption{Model-Based Entropy-Regularized Imitation Learning (MB-ERIL)}
\label{alg:MB-ERIL}
\begin{algorithmic}[1]
  \Require{Expert dataset $\mathcal{D}^E$ and hyperparameters
    $\kappa$ and $\eta$}
  \Ensure{Learner's policy $b$, model $q$, reward $r$,
    state-value $V$, and state-action value $Q$.} 
  \State Initialize all parameters of networks and replay buffers
  $\mathcal{D}^L, \mathcal{D}^G$.
  \State (Pre-train the Regularized AutoEncoder using $\mathcal{D}^E$
  when raw images are used as inputs.)
  \For{$k = 0, 1, 2, \ldots$}
  \LeftComment {Collect state transitions in the real
  environment and the model.}
  \State $\mathcal{D}^L \leftarrow \mathcal{D}^L \cup
  \{ (\bx_t, \bu_t, \bx_{t+1} ) \}_{t=0}^{N^L}$ with 
  $b$ and $p$.
  \State $\mathcal{D}^G \leftarrow \mathcal{D}^G \cup
  \{ (\bx_t, \bu_t, \bx_{t+1} ) \}_{t=0}^{N^G}$ with 
  $b$ and $q$.
  \LeftComment{Train the discriminators}
  \State Update $r$, $V$, and $Q$ by minimizing 
  \eqref{eq:loss_D}.
  \LeftComment{Evaluate policy $\qquad \hspace*{43.5mm}.$}
  \State $\mathcal{D}^G \leftarrow \mathcal{D}^G \cup
  \{ (\bx_t, \bu_t, \bx_{t+1} ) \}_{t=0}^{N^G}$ with 
  $b$ and $q$.
  \State Update $V$ and $Q$ by minimizing \eqref{eq:VQ_update_rule}.
  \LeftComment{Improve the policy and the model $\qquad \hspace*{27.8mm}.$}
  \State Update $q$ and $b$. See
    \eqref{eq:loss_model} and  \eqref{eq:loss_policy}.
  \EndFor
\end{algorithmic}
\end{algorithm}

\subsection{Simplified MB-ERIL}
\label{sec:simplified_MB-ERIL}

Here we show two simplified MB-ERIL implementations by removing
some components.
One is MB-ERIL without training the discriminators, which
trains $r$, $V$, and $Q$ by the maximum likelihood method
instead of minimizing \eqref{eq:loss_D}.
The loss function is given by
\begin{align}
  &
 \mathcal{L}' (r, V, Q) = - \mathbb{E}_{\mathcal{D}^E}
 \left[ \ln p (\bx' \mid \bx, \bu) \right]
 \notag \\
  &\quad 
  - \mathbb{E}_{\mathcal{D}^E} \left[ \ln b(\bu \mid \bx) \right]
  + \mathcal{L} (V, Q).
 \label{eq:MB-ERIL_wo_D}
\end{align}
The first term of the right-hand side of (1) corresponds to
model cloning, and the third term is the regularizer, which resembles
SQIL.
We call this approach Entropy-Regularized Model and Behavior
Cloning (ERMBC). 

The other is MB-ERIL without training the generator inspired
by ASAF. This approach simply skips lines 7 and 8 of
Algorithm~\ref{alg:MB-ERIL}, 
meaning that it does not learn from the additional simulated
trajectories even though the model is maintained explicitly.
We call this approach MB-ERIL without Policy Evaluation
(MB-ERIL{\textbackslash}PE).

\section{MUJOCO BENCHMARK CONTROL TASKS}
\label{sec:mujoco}

\subsection{Task description}

To verify our proposed method, we first conducted two benchmark
control tasks, Ant and Humanoid, provided by OpenAI gym
\cite{Brockman2016a}.
Both use a physics engine called MuJoCo \cite{Todorov2012a}.
The goal of these tasks is to move forward as quickly as possible.
First, an optimal policy was trained by the Trust Region Policy
Optimization \cite{Schulman2015a}, based on the original reward
function provided by the simulator.
Then it was used as an expert policy to collect
expert data $\mathcal{D}^E$.
We adopted the sparse sampling setup
\cite{Sasaki2019a} in which we randomly sampled $(\bx, \bu, \bx')$
triplets from each trajectory.

The functions used by MB-ERIL need to be approximated by artificial
neural networks, and $r(\bx)$, $V(\bx)$, $Q(\bx, \bu)$, and
$b(\bu \mid \bx)$ were represented by a two-layer neural network
with a derivative of the sigmoid-weighted linear unit (dSiLU)
\cite{Elfwing2018a} as an activation function, 
determined based on our previous study \cite{Uchibe2021a}. 
We defined the model to output a Gaussian distribution
conditioned on the state and the action,i.e.:
$q(\bx' \mid \bx, \bu) = \mathcal{N}(\bx' \mid \bm{\mu}
(\bx, \bu), \bm{\Sigma}(\bx, \bu))$, where $\bm{\mu}(\bx, \bu)$
and $\bm{\Sigma}(\bx, \bu)$ denote the mean vector and the diagonal
covariance matrix 
for simplicity. Future work will investigate more complicated distributions,
such as a mixture of Gaussians with full covariance.
Each was represented by a neural network with two hidden
layers, 
and each layer had 256 nodes
with dSiLU activations.

\subsection{Comparative evaluation}
\label{sec:mujoco:results}

We evaluated the MB-ERIL, ERMBC, and
MB-ERIL{\textbackslash}PE performances by comparing them with the following
algorithms:
\begin{itemize}
\item Dyna-MF-ERIL: MF-ERIL \cite{Uchibe2021a} with the
  Dyna framework \cite{Sutton1990a}, which generates
  simulated trajectories by interacting with the model. 
  The model, represented by a separate neural network, is
  trained by the simple maximum likelihood method.
  Then MF-ERIL is trained from real and generated data.
\item MF-ERIL: We ignored the first discriminator that
  estimates $\ln p_0^E(\bx) / p_0^L(\bx)$ for simplicity.
\item DAC: Discriminator-Actor-Critic (DAC)
  \cite{Kostrikov2019a}, which is baseline model-free
  imitation learning.
\item BC: naive Behavior Cloning that minimizes the negative
  log-likelihood objective.
\end{itemize}
Following Ho and Ermon \cite{Ho2016c}, a single
trajectory contains 50 state-action transition pairs
$(\bx, \bu, \bx')$, i.e., 50 steps per episode.
The number of trajectories sampled from expert policy
$\pi$ were set to 30 and 350 in the Ant and
Humanoid environments. 
We set $N^L = 10^2$ and $N^G = 10^4$.

\begin{figure}[t]
  \centering
  \begin{tabular}{c}
    \includegraphics[width=1.0\linewidth]{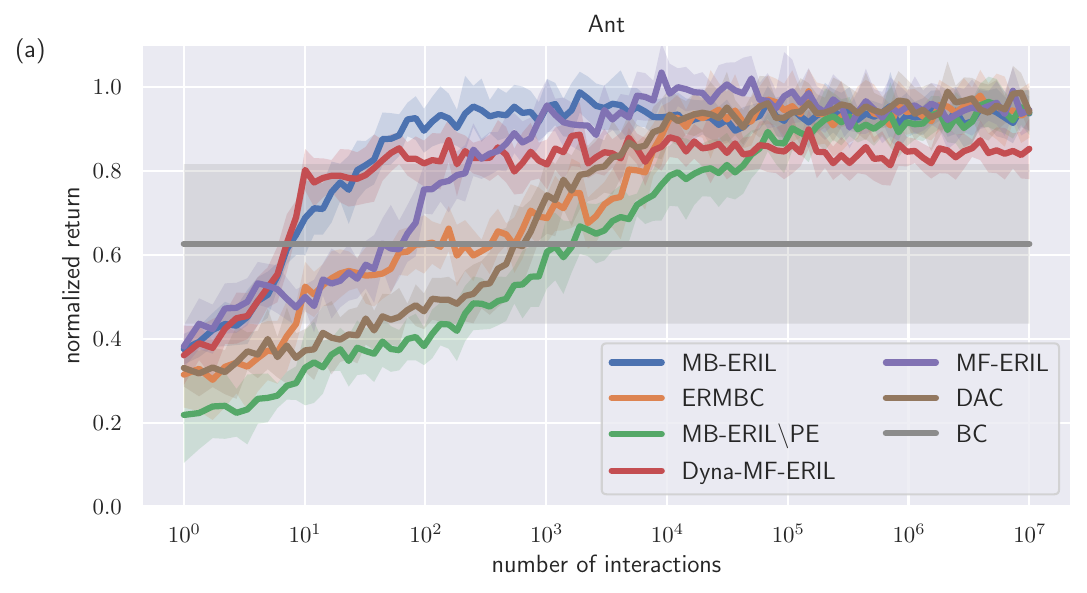}
    \\
    \includegraphics[width=1.0\linewidth]{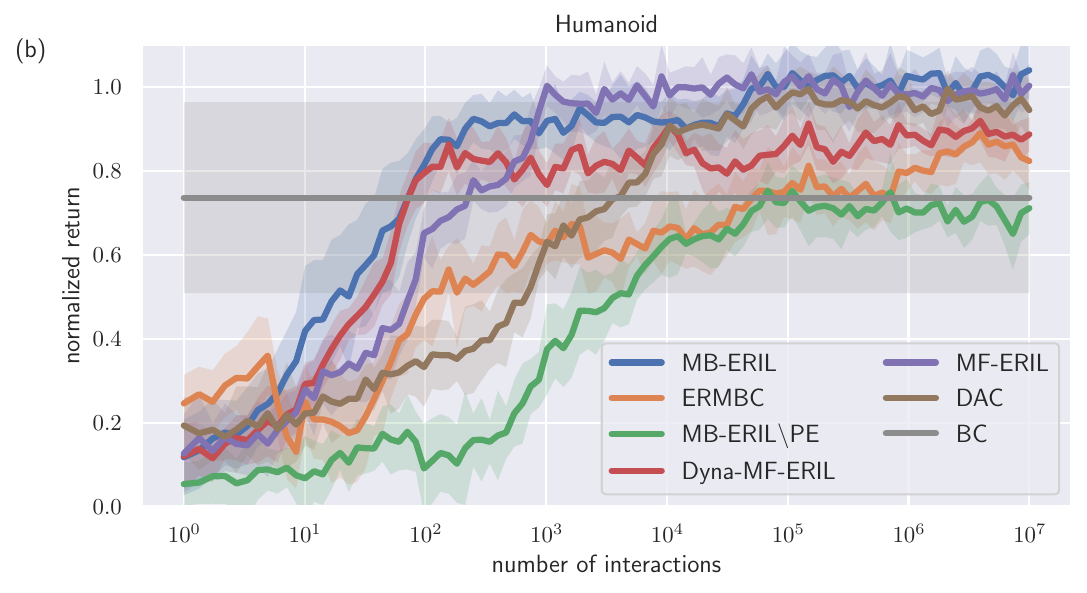} 
  \end{tabular}
  \caption{Normalized return versus number of interactions
    on MuJoCo benchmark control tasks:
    Solid lines represent average values, and shaded areas
    correspond to $\pm 1$ standard deviation region.
    Note that horizontal axis is a log scale.}
    \label{fig:mujoco:frl:normal}
\end{figure}

Fig.~\ref{fig:mujoco:frl:normal} compares the learning curves\footnote{
We conducted Hopper, Walker, Reacher, and HalfCheetah, all of which we previously
evaluated \cite{Uchibe2021a}.
Simulation results show that MB-ERIL outperformed MF-ERIL,
although we omitted them due to space limitations.}.
Note that the reward functions estimated by each method
cannot be directly compared since inverse reinforcement
learning is an ill-posed problem.
Therefore, the method was evaluated by the mean normalized
return:
\begin{displaymath}
  \bar{R}_{\mathrm{mean}} = \frac{1}{N}
  \sum_{i=1}^N \frac{R_i - R_{\min}}{R_{\max} - R_{\min}},
\end{displaymath}
where $i$ is the index of the experimental run and $N$ is
the number of experiments.
$R_i$ is the raw return of the $i$-th experimental run,
where the original reward of OpenAI gym was used.
$R_{\min}$  and $R_{\max}$  are constants, where $R_{\max}$ is
given by the return of the expert while $R_{\min} = 0$.
In the Ant environment, the maximum normalized total
reward was reached by MB-ERIL, MF-ERIL, DAC, ERMBC, and
MB-ERIL{\textbackslash}PE in that order;
Dyna-MF-ERIL and BC did not reach the expert performance.
MB-ERIL improved the learning efficiency of the model-free
methods (MF-ERIL and DAC) by about ten times.
Note that the asymptotic performance of Dyna-MF-ERIL was
worse than that of MF-ERIL even though the performing model
did learn and its normalized return increased rapidly
in the early stage of learning.
Perhaps the model failed to learn the
state transition because model learning by the maximum
likelihood method cannot deal with a covariate shift.
ERMBC and MB-ERIL{\textbackslash}PE also converged to the
expert performance, although they learned much slower than
MF-ERIL. 
Similar results were obtained in the Humanoid environment.
MB-ERIL learned faster than the other methods, although
MF-ERIL and DAC eventually achieved comparable performance.
The performance of MB-ERIL{\textbackslash}PE
was worst or inferior to BC.

\begin{figure}[t]
  \centering
  \begin{tabular}{c}
    \includegraphics[width=1.0\linewidth]{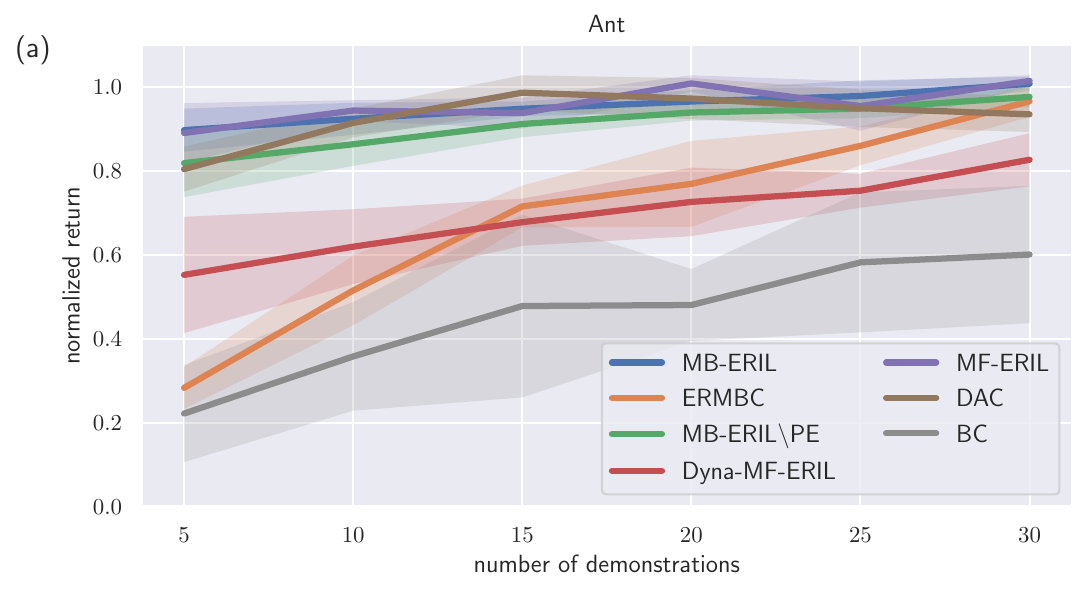}
    \\
    \includegraphics[width=1.0\linewidth]{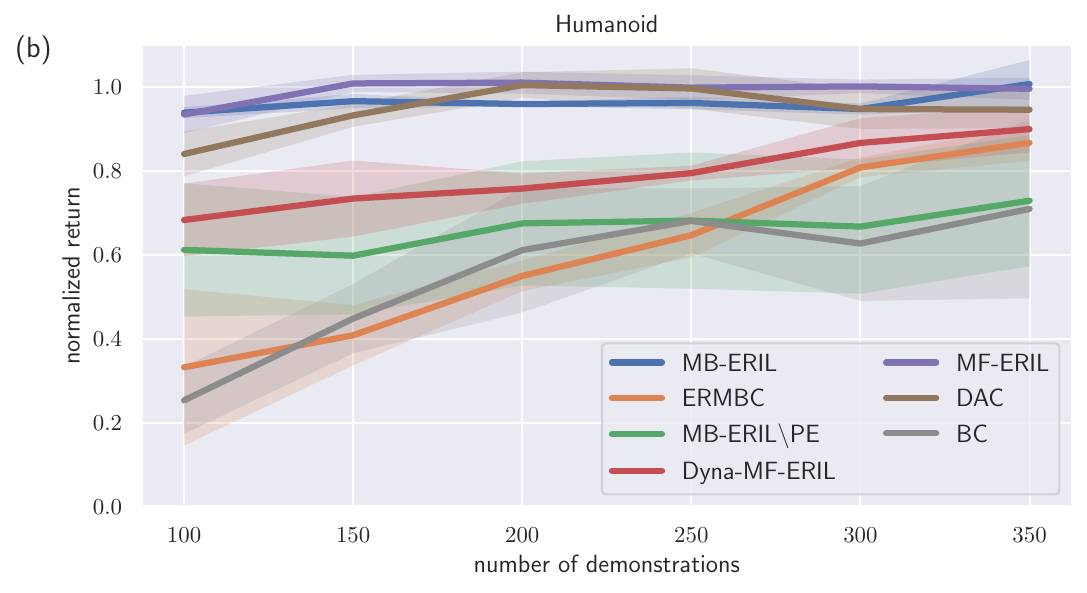} 
  \end{tabular}
  \caption{Normalized return versus number of trajectories
    provided by expert on MuJoCo benchmark control tasks}
  \label{fig:mujoco:irl:normal}
\end{figure}

Next we evaluated the data efficiency of the training
discriminators by changing the number of samples
in $\mathcal{D}^E$. 
Fig.~\ref{fig:mujoco:irl:normal} shows that
MB-ERIL, MB-ERIL{\textbackslash}PE, MF-ERIL, and DAC
found policies that exhibited higher control performance with
fewer expert demonstrations. 
On the other hand, ERMBC and Dyna-MF-ERIL just exhibited slight 
performance degradation when the number of demonstrations was
limited, implying that model learning did
not improve the sample efficiency with respect to the number of
expert demonstrations.
However, the  MB-ERIl{\textbackslash}PE performance was
degraded in the Humanoid environment; it
did not improve even though the number
of demonstrations increased.
The results indicate that $V$ and $Q$ did not satisfy
the soft Bellman equations \eqref{eq:V_and_Q1} and
\eqref{eq:V_and_Q2} when they were simply estimated by
training the discriminators. 
We conclude that the policy evaluation of MB-ERIL
was more crucial when the state space is large.

\begin{figure}[t]
  \centering
  \includegraphics[width=1.0\linewidth]{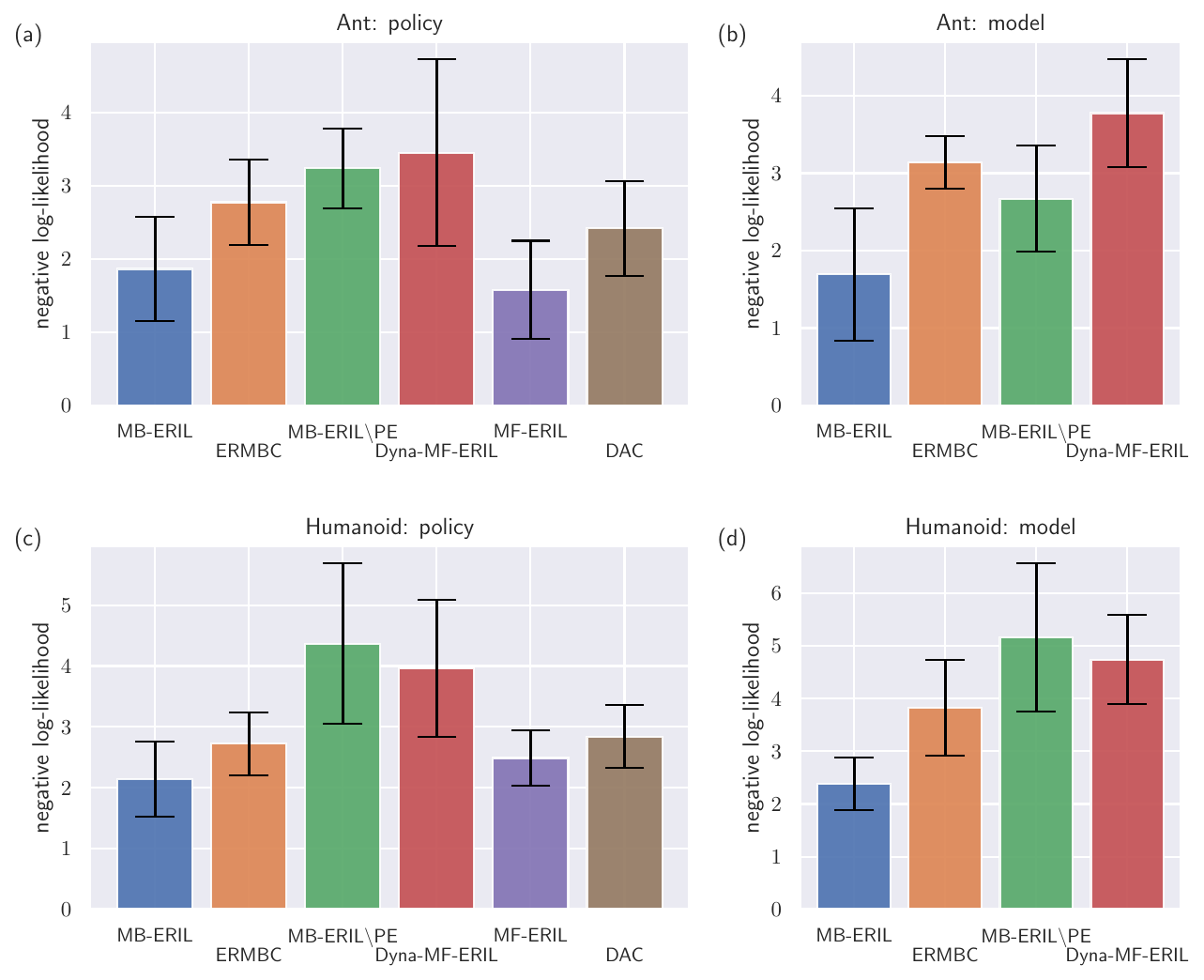}
  \caption{Comparison of NLL in Ant and Humanoid environments:
    Smaller is better because vertical axis is negative
    log-likelihood. 
    (a) and (c) NLL of policy.
    (b) and (d) NLL of model. Note that MF-ERIL
    and DAC did not estimate the model.}
  \label{fig:Humanoid:frl:NLL}
\end{figure}

To see the model and policy differences among MB-ERIL,
Dyna-MF-ERIL, and MF-ERIL, we computed the negative
log-likelihood (NLL) of the finally-obtained model and
policy using test expert data.
For example, the model's NLL was calculated by
\begin{displaymath}
  \mathrm{NLL} = - \frac{1}{N_{\mathrm{test}}^E} 
  \sum_{\ell=1}^{N_{\mathrm{test}}^E} \ln q (\bx_\ell' \mid
  \bm{x}_\ell, \bu_\ell),
\end{displaymath}
where $N_{\mathrm{test}}^E$ is the number of samples in the
test data.
The policy's NLL was defined similarly.
Figs.~\ref{fig:Humanoid:frl:NLL}(a) and (c) compare the NLLs of the
policies on the MuJoCo experiments, showing 
no significant difference between the MB-ERIL and MF-ERIL policies, although 
Dyna-MF-ERIL's NLL was larger than the others, which led to its
poor asymptotic performance.
Figs.~\ref{fig:Humanoid:frl:NLL}(b) and (d) show that the NLL of the
model obtained by MB-ERIL was the smallest among the
other methods.

\section{REAL ROBOT CONTROL}
\label{sec:nextage}

\subsection{Task description}
\label{sec:nextage:setting}

\begin{figure}[t]
  \centering
  \includegraphics[width=1.0\hsize]{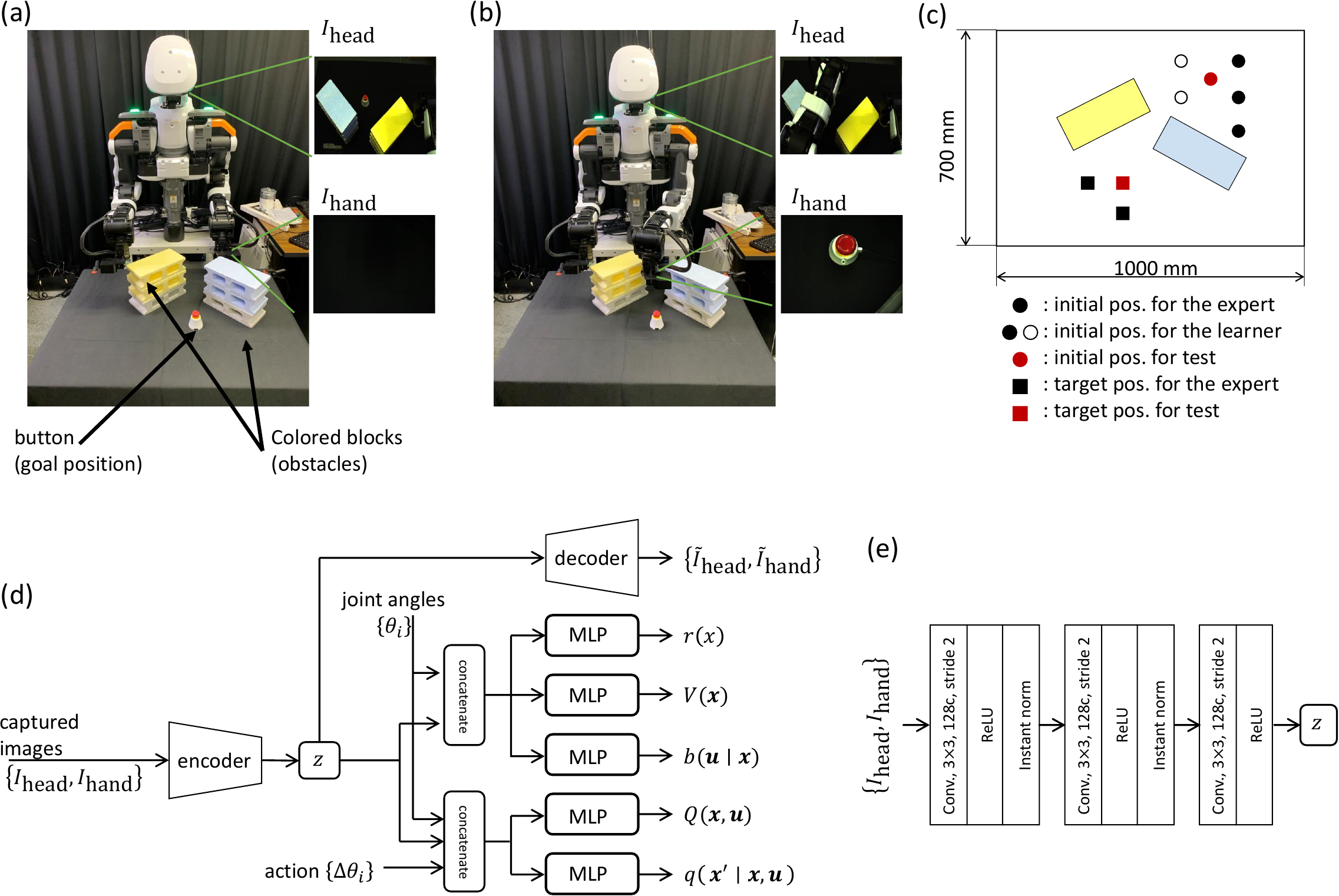}
  \caption{Vision-based reaching task: (a) Starting position.
    (b) Target position. (c) Possible environmental configuration.
    (d) Network architecture for reward, state value, policy,
    state-action value, and model. (e) Network architecture for
    encoder: Conv. denotes a convolutional neural network.
    ``$n$c'' denotes ``$n$ channels.''}
  \label{fig:nextage}
\end{figure}

Next we performed a vision-based reaching task \cite{Uchibe2021a}
with an upper-body humanoid called Nextage developed by Kawada
Industries Inc.
Two colored blocks served as obstacles, and one switch button
indicated the goal in the workspace.
The aim is to move its left arm's end-effector from
a starting position (Fig.~\ref{fig:nextage}(a)) to a target
position (Fig.~\ref{fig:nextage}(b)) as quickly as possible,
where the starting and target positions were selected randomly
from the pre-defined positions in Fig.~\ref{fig:nextage}(c).
This task is more complicated than the FetchReach provided by
OpenAI Gym because the target position must be detected from
visual information. 
Nextage has a head with two cameras, a torso, two 6-axis
manipulators, and two cameras attached to its end-effectors.
We mounted cameras on its left arm and head in this task.
The head pose was fixed during the experiments.

The state was given by
$\bx = [\theta_1, \ldots, \theta_6, \bm{z}^\top]^\top$, where
$\theta_i$ is the $i$-th joint angle of the left arm and
$\bm{z}$ represents the latent visual state obtained by a
deterministic Regularized AutoEncoder (RAE) \cite{Ghosh2020a}.
The action was given by the changes in the joint angle from
previous position
$\bu = [\Delta \theta_1, \ldots, \Delta \theta_6]^\top$, where
$\Delta \theta_i$ is the change in the joint angle from the
previous position of the $i$-th joint (although we clamped
$\Delta \theta_i$ in $[-1, 1]$. 
Figs.~\ref{fig:nextage}(d) and (e) show the entire network
architecture that represents the functions and the RAE encoder
whose input came from two $160 \times 128$ RGB images
captured by Nextage's cameras.

We prepared several environmental configurations by changing the arm's
initial pose, the button's location, and the height of the blocks. We
designed three initial and two target poses for the expert
configuration (Fig.5(c)). The learning configuration was given
by five initial and two target poses. In addition, we prepared blocks
of two different heights, resulting in $3 \times 2 \times 2 = 12$ expert
configurations and $5\times 2 \times 2 = 20$ learning configurations.
We created expert
demonstrations with MoveIt! \cite{Chitta2012a} using the geometric information of the button
and the colored blocks. Note that such information was 
unavailable for learning the algorithms.
MoveIt! generated 12 trajectories for every expert configuration,
and we sampled 50 transitions for each trajectory.
Consequently, $N^E = 12 \times 50$ expert transitions were
obtained. Although the sequence of the joint angles was almost
deterministic, the RGB images were stochastic and noisy due to
the lighting conditions.
In this experiment, $N^L$ and $N^G$ were set to
$20 \times 50$ and $10^4$.
On the other hand, the test configuration was constructed by one initial
pose (red circle) and one target one (red square) that were
not included in the expert and learner's configurations.

\subsection{Experimental results}
\label{sec:nextage:results}

We compared MB-ERIL with ERMBC, MB-ERIL{\textbackslash}PE,
Dyna-MF-ERIL, MF-ERIL, DAC, ASAF, and BC.
We measured the performances by investigating a
synthetic reward function:
$r = \exp \left( - \frac{\| \bm{p}_t - \bm{p}_g \|_2^2}{\sigma^2}
\right)$,
where $\bm{p}_t$ and $\bm{p}_g$ denote the end-effector's current
and target position.
Parameter $\sigma$ was set to 5.

To evaluate the performance during training, we evaluated the policy's
performance every ten episodes with a test set
of initial and goal positions that were not included in the
training demonstrations from the expert.

\begin{figure}[t]
  \centering
  \includegraphics[width=1.0\hsize]{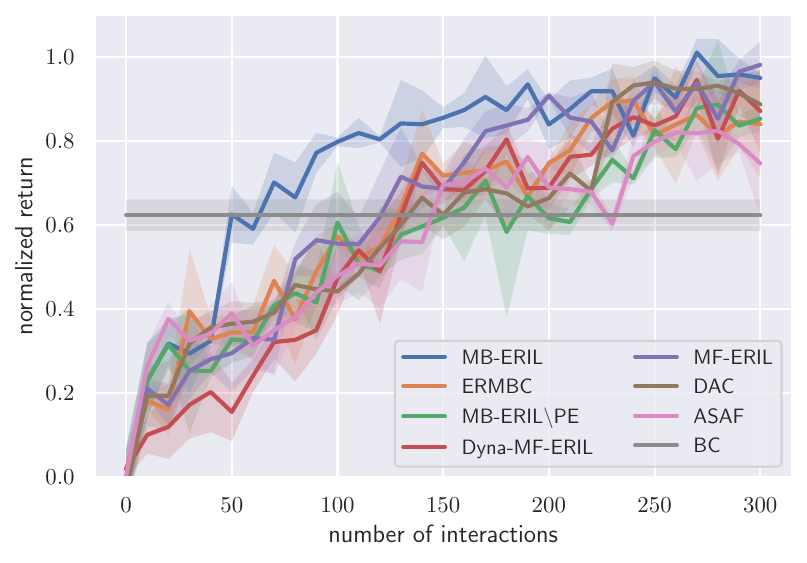}
  \caption{Normalized return versus number of interactions
    on Nextage vision-based reaching task}
  \label{fig:nextage:frl:normal}
\end{figure}

\begin{figure}[t]
  \centering
  \includegraphics[width=1.0\linewidth]{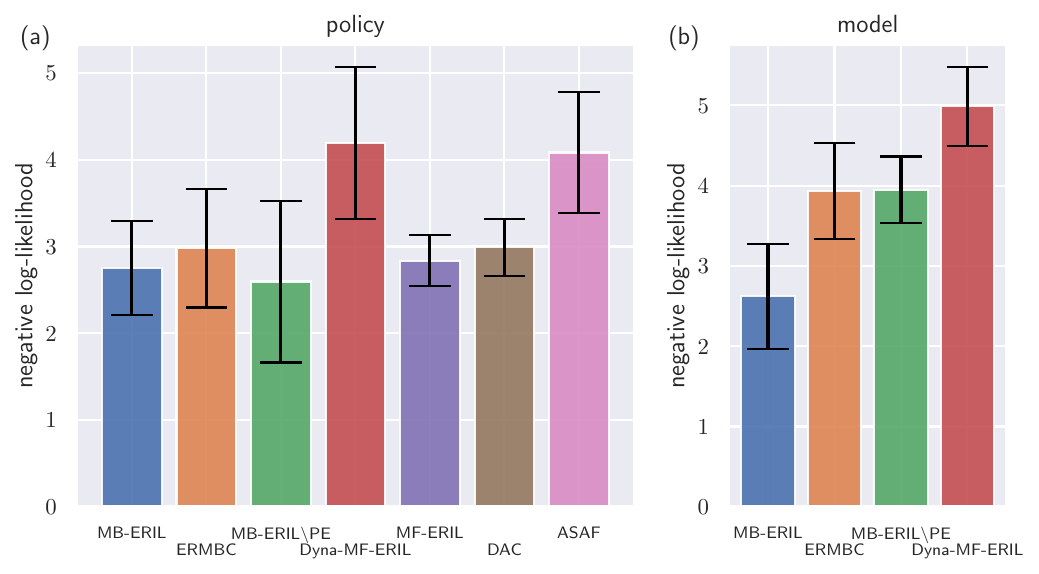}
  \caption{Comparison of NLL on Nextage vision-based
    reaching task: (a) Policy's NLL and (b) Model's NLL}
  \label{fig:nextage:NLL}
\end{figure}

Fig.~\ref{fig:nextage:frl:normal} compares the performance,
where the horizontal and vertical axes represent the number of
environmental interactions and normalized returns.
MB-ERIL achieved higher sample efficiency than the other baselines.
Unlike the MuJoCo experiments, Dyna-MF-ERIL was comparable
to MF-ERIL, indicating that the model bias was not serious in
this task.
However, Dyna-MF-ERIL did not exploit the model efficiently
because it did not improve the sample efficiency.
Fig.~\ref{fig:nextage:NLL} shows that the NLL of MB-ERIL's
policy was comparable to those of ERMBC,
MB-ERIL{\textbackslash}PE, MF-ERIL, and DAC, although the
NLL of MB-ERIL's model was the smallest.

\section{CONCLUSIONS}
\label{sec:conclusions}

We proposed MB-ERIL, derived from RL theory, which
regularized the policy and the model.
The experimental results of the simulated and real robots showed that
MB-ERIL improved the sample efficiency over MF-ERIL, Dyna-MF-ERIL,
DAC, and BC.
A comparison between MB-ERIL and Dyna-MF-ERIL suggests that the simple
use of data generated by the model was not useful due to the
model bias.   \revised{
    As explained in Section~\ref{sec:introduction}, our formulation is based on the
    assumption that the expert policy and the actual environment
    are solutions of the regularized Bellman equation, although
    this assumption is not straightforward.
    In the future, we will provide a clear derivation of ERIL.
  }{black}

Our experimental results on the Ant task show that MB-ERIL's
asymptotic performance was slightly worse than that of MF-ERIL
due to model bias.
One possible solution is to switch the algorithm from MB-ERIL to
MF-ERIL based on
the loss function of the model discriminator.
MB-ERIL updates $V$ and $Q$ alternatively with respect to two different
objective functions. Although the update rules were derived from the
Bellman equation \eqref{eq:Bellman_equation}, the convergence and
stability were not proved theoretically.
In future work, we will apply the technique used in the convergence
proof of GAIL \cite{Zhang2020a, Guan2021a}.

\addtolength{\textheight}{-4mm}  % This command serves to balance the column lengths

%%%%%%%%%%%%%%%%%%%%%%%%%%%%%%%%%%%%%%%%%%%%%%%%%%%%%%%%%%%%%%%%%%%%%%%%%%%%%%%%
\appendices
\section*{APPENDIX}

\subsection{Derivation of \eqref{eq:model_update_rule} -
\eqref{eq:V_and_Q2}}
\label{sec:derivation:Bellman}

Consider the maximization problem inside the right-hand
side of \eqref{eq:Bellman_equation}. It is a constrained
optimization because $\int p(\bx' \mid \bx, \bu)\mathrm{d}\bx' = 1$.
The Lagrangian is given by
\begin{multline}
  \mathcal{L}_2 = \int p \left( r - \kappa^{-1}p - \eta^{-1}
    \ln \frac{p}{q} + \gamma V' \right) \mathrm{d}\bx' \\
  + \lambda_1 \left(1 - \int p \mathrm{d}\bx' \right),
  \notag
\end{multline}
where $\lambda_1$ is the Lagrangian multipliers.
Note that we simplified the notation here to improve its
readability. We set the derivative with respect to $p$ to 0
and used the
constraint to yield the following equation:
\begin{displaymath}
  p = \frac{\exp [\beta (r + \gamma V' + \eta^{-1} \ln q)]}
  {\exp (1 + \beta \lambda_1)},
\end{displaymath}
where the denominator is given by
\begin{equation}
  \exp (1 + \beta \lambda_1) = \int \exp [\beta
  (r + \gamma V' + \eta^{-1} \ln q) \mathrm{d}\bx'.
  \label{eq:solution:Lagrange1}
\end{equation}
By defining the right-hand side of \eqref{eq:solution:Lagrange1}
as $\exp \beta Q$, we obtain \eqref{eq:model_update_rule} and
\eqref{eq:V_and_Q1}.
Substituting the above results into \eqref{eq:Bellman_equation}
yields
\begin{displaymath}
  V = \max_{\pi} \mathbb{E}_{\pi} \left[ -\kappa^{-1} \ln \pi
    - \eta^{-1} \ln \frac{\pi}{b} + Q \right].
\end{displaymath}
Similarly, we can maximize the right-hand side using the
Lagrangian multipliers method and
obtain \eqref{eq:policy_update_rule} and
\eqref{eq:V_and_Q2}.

% \section*{ACKNOWLEDGMENT}

%%%%%%%%%%%%%%%%%%%%%%%%%%%%%%%%%%%%%%%%%%%%%%%%%%%%%%%%%%%%%%%%%%%%%%%%%%%%%%%%

\bibliographystyle{IEEEtran}
\bibliography{reference}

\end{document}